\documentclass{article}
\usepackage{graphicx} 
\usepackage{hyperref}
\usepackage{amsmath}
\usepackage{float}
\usepackage{bbm}
\usepackage{outlines}
\usepackage{subcaption}
\usepackage{amssymb}
\usepackage{xcolor}

\usepackage[ruled,vlined]{algorithm2e}

\usepackage[a4paper, total={7in, 9in}]{geometry}

\title{Balancing Accuracy and Speed: A Multi-Fidelity Ensemble Kalman Filter with a Machine Learning Surrogate Model}

\author{Jeffrey van der Voort \footnote{Delft Institute of Applied Mathematics, Delft University of Technology, Delft, The Netherlands;\\ email: j.c.vandervoort@tudelft.nl}  , Martin Verlaan \footnote{Delft Institute of Applied Mathematics, Delft University of Technology, Delft, The Netherlands} \footnote{Deltares, Delft, The Netherlands} \ and Hanne Kekkonen \footnote{Delft Institute of Applied Mathematics, Delft University of Technology, Delft, The Netherlands}}

\date{December 13, 2025}

\begin{document}

\maketitle

\begin{abstract}
Currently, more and more machine learning (ML) surrogates are being developed for computationally expensive physical models. In this work we investigate the use of a Multi-Fidelity Ensemble Kalman Filter (MF-EnKF) in which the low-fidelity model is such a machine learning surrogate model, instead of a traditional low-resolution or reduced-order model. The idea behind this is to use an ensemble of a few expensive full model runs, together with an ensemble of many cheap but less accurate ML model runs. In this way we hope to reach increased accuracy within the same computational budget. We investigate the performance by testing the approach on two common test problems, namely the Lorenz-2005 model and the Quasi-Geostrophic model. By keeping the original physical model in place, we obtain a higher accuracy than when we completely replace it by the ML model. Furthermore, the MF-EnKF reaches improved accuracy within the same computational budget. The ML surrogate has similar or improved accuracy  compared to the low-resolution one, but it can provide a larger speed-up. Our method contributes to increasing the effective ensemble size in the EnKF, which improves the estimation of the initial condition and hence accuracy of the predictions in fields such as meteorology and oceanography. 
\end{abstract}


\section{Introduction}
Data assimilation is widely applied in meteorology (\cite{zhang2009}; \cite{houtekamer2016}), oceanography (\cite{yin2011}; \cite{ohishi2023}), and many other fields (e.g., \cite{reichle2002}; \cite{zhou2006}). The physical models used in these fields are often computationally expensive, which limits the affordable ensemble size for the popular Ensemble Kalman Filter (EnKF) \cite{evensen1994}. Several approaches have been proposed to reduce the computational cost of these models, making use of a cheap proxy or surrogate of the expensive physical model. \\
\ \\
Traditionally, a low-resolution (LR) or reduced-order model (ROM) surrogate is used to increase the ensemble size within the same computational budget. One way to include the surrogate information is by using a multi-level or multi-fidelity approach, see \cite{giles2015} and \cite{peherstorfer2018} respectively. The idea behind these approaches is to use an ensemble of a few computationally expensive high-resolution (HR) runs, combined with many cheap and less accurate surrogate runs. In this way one can reach similar or improved accuracy compared to the HR model, while reducing computational time. For instance, \cite{popov2021} apply a multi-fidelity EnKF (MF-EnKF) with a ROM surrogate to a quasi-geostrophic (QG) model and show that it is more stable than an optimally tuned EnKF for the same number of HR runs. Further, \cite{hoel2016} use the multi-level approach, called multi-level EnKF (ML-EnKF), where on each level there is a pair-wise coupling between the HR and LR ensemble, and show that it outperforms the standard EnKF in two common stochastic differential equation problems. In fact, one can show that the MF-EnKF is a generalization of the ML-EnKF, but with a correction factor for limited correlation \cite{popov2021}. However, the disadvantage of the multi-level approach is that the background covariance is not guaranteed to be symmetric positive definite \cite{giles2015}. \\
\ \\
Besides multi-level and multi-fidelity, there are other approaches to combine the ensembles. For instance, in the Mixed-Resolution EnKF by \cite{rainwater2013}, a HR ensemble is combined with a LR ensemble by using a hybrid background covariance matrix in the analysis step. Another approach is the Super-Resolution EnKF by \cite{barthelemy2022}, in which the forecast step only uses a LR ensemble. Then in the analysis step the LR ensemble is downscaled to HR space and assimilated with HR observations. \cite{barthelemy2024} have later also combined the Super-Resolution EnKF with the Mixed-Resolution EnKF. \\
\ \\
The disadvantage of using LR or ROM surrogates is that they can have limited speed-ups. Furthermore, unresolved scales or sub-grid-scale processes, such as the cloud physics in a global atmospheric model, are often inaccurate and limit the predictive performance of these models. Also, LR models can suffer from representation errors, since sharp gradients, such as sea-land transition, cannot be represented well on the LR grid. \\
\ \\
Recently, Machine Learning (ML) methods have been proposed as a surrogate model. In Table \ref{tab:ml_weather} we give an overview of some of the ML surrogates developed in the field of weather prediction and their speed-up factors. We should note that some of the speed-up factors also take into account the number of CPU/GPU nodes and hence they are even larger than for the other ML surrogates. Typically, these ML models are very fast since they can get a speed up by performing computations on the GPU. This makes the speed-up factors not entirely fair, since most traditional numerical weather prediction models, such as the IFS \cite{isaksen2010}, perform computations on the CPU. Another advantage of ML surrogates is that they can be trained on the HR model output or directly on observations, thus avoiding inaccurate representations or unresolved scales.

\begin{table}[H]
    \centering
      \caption{ML surrogates for weather prediction and their speed-up factors compared to numerical weather prediction. Speed-up is compared to IFS, which takes 6 hours to produce the forecast. Some of the speed-ups also include the number of CPU/GPU nodes required next to computational time.}
    \begin{tabular}{c|c}
       \textbf{ML surrogate}  & \textbf{Speed-up} \\ 
       \hline
        Aardvark \cite{aardvark} & $>21,600$ \\
        AIFS \cite{aifs} & 144 \\
        Aurora \cite{aurora} & Several orders of magnitude \\
        FourcastNet \cite{fourcastnet} & 80,000 \\
        Graphcast \cite{graphcast} & $>360$ \\
        PanguWeather \cite{pangu} & 10,000 \\
    \end{tabular}
    \label{tab:ml_weather}
\end{table}

The most common approach to include ML is as a full replacement for the physical model. For example, \cite{maulik2022} and \cite{melinc2024} use a ML surrogate inside variational data assimilation, so that automatic differentiation can be used to obtain the adjoint in a cheap way. There are also some approaches that can learn from noisy and sparse observations (e.g., \cite{brajard2020}; \cite{bocquet2020}) by alternating a ML step with a DA step. Furthermore, ML can be used to learn the model error \cite{farchi2025} or background covariance \cite{sacco2024}. Alternatively, in the approach by \cite{vikram2022}, a HR and ML ensemble are propagated separately. Then in the analysis step, the ML ensemble is recentered using the mean of the HR ensemble and then merged into one ensemble. For an overview of methods that combine ML and DA, we refer the reader to the reviews by \cite{buizza2021} and \cite{cheng2023}. \\
\ \\
On the other hand, ML surrogates are usually less accurate than the high-resolution model. For instance, \cite{bonavita2024} shows that the recently developed ML surrogates for weather prediction all underestimate the vertical wind velocity, which is crucial for typhoon prediction, even though the track of the typhoon is predicted very well. Furthermore, \cite{zhang2025} show that ML models underperform in the prediction of extreme weather events. Also in combination with DA, ML surrogates can have issues. \cite{slivinski2025} show that the same ML surrogates as mentioned earlier become unstable when combined with DA for the prediction of 500hPa geopotential height. \\
\ \\
This paper will combine the multi-fidelity approach with a ML surrogate model. We use the computational speed of the ML surrogate, and the stability and accuracy provided by the multi-fidelity approach. In particular, we investigate the use of a Multi-Fidelity Ensemble Kalman Filter (MF-EnKF) \cite{popov2021} in which the low-fidelity surrogate is a ML model instead of a traditional LR or ROM one. We investigate the performance by testing the approach on two common test problems, namely the Lorenz-2005 model \cite{lorenz2005} and the Quasi-Geostrophic model \cite{thiry2024}. By keeping the original physical model in place, we obtain higher accuracy than when we completely replace it by the ML model, as is often done in surrogate modeling. Also for a fixed computational budget, the MF-EnKF reaches improved accuracy compared to the EnKF. Finally, the ML surrogate achieves comparable accuracy but with a faster speed compared to a traditional LR surrogate. \\
\ \\
We should note that the MF-EnKF has been combined with ML before. \cite{popov2022ae} combine the MF-EnKF with an Auto-Encoder (AE), which learns the mapping from HR to ROM space. It is shown that the AE can create a more accurate ROM than projection-based methods. The contribution of our approach is that we instead learn a ML model for the forward mapping, which operates in the same HR space as the physical model. This removes the need to map between the HR and LR spaces, and thus avoids the inaccuracies introduced there. Our approach is general and can in principle be used for all EnKF applications. It is especially attractive whenever a high quality ML model is already available, such as PanguWeather \cite{pangu}, Graphcast \cite{graphcast} and FourCastNet \cite{fourcastnet} in the field of numerical weather prediction.


\section{Theoretical background} \label{theory}

\subsection{Ensemble Kalman Filter (EnKF)} \label{enkf}
First, we introduce the notation and the standard Ensemble Kalman Filter (EnKF) \cite{evensen1994}. We assume the physical model satisfies a state equation of the following form:
\begin{equation} \label{eq:state_eq}
    X_k = \mathcal{M}_{X} (X_{k-1} ) + \varepsilon_k,
\end{equation}
where $X_k \in \mathbb{R}^n$ denotes the state vector at discrete time $t_k$, $\mathcal{M}_{X}$ the forward model, which is the discrete time representation of the physical model that maps the previous state to the next one, and $\varepsilon_k \sim \mathcal{N}(0,Q_k)$ the model error. In this paper we assume there is no model error term ($Q_k = O)$, since we assume that the available physical model is the true model. \\
\ \\
In the forecast step of the EnKF, we propagate each ensemble member forward in time using the state equation \eqref{eq:state_eq}:
\begin{equation} \label{eq:enkf_for}
    X_k^{b, (i)} = \mathcal{M}_X(X_{k-1}^{a, (i)} ), \qquad \qquad i = 1, \ldots, N_X, 
\end{equation} 
where $X_{k-1}^{a,(i)} \in \mathbb{R}^n$ is the analysis at previous time $t_{k-1}$ and $X_k^{b,(i)}$ the forecast or background at time $t_k$. Running the model $\mathcal{M}_X$ in the forecast step is often the most expensive part of the EnKF. We denote by $E_{X_k^b}$ the matrix containing all ensemble members $X_k^{b,(i)} \in \mathbb{R}^n$ as its columns: 
$$ E_{X_k^b} = \begin{bmatrix}
    X_k^{b,(1)} \cdots X_k^{b, (N_X)}
\end{bmatrix} \in \mathbb{R}^{n \times N_X}. $$
Furthermore, we define the ensemble anomalies as
$$ A_{X_k^b} = \frac{1}{\sqrt{N_X-1} } (E_{X_k^b} - \mu_{X_k^b} ), $$
where $\mu_{X_k^b}$ is the mean of the ensemble $E_{X_k^b}$. The prior Gaussian distribution at time $t_k$ is characterized by the ensemble mean $\mu_{X_k^b}$ and the ensemble covariance $\Sigma_{X_k^b} = A_{X_k^b} A_{X_k^b}^T$. \\
\ \\
Furthermore, we assume that the observations are generated according to the observation equation, given by
\begin{equation} \label{eq:obs_eq}
    y_k = \mathcal{H} (X_k) + \eta_k,
\end{equation}
where $y_k \in \mathbb{R}^m$ denotes the vector of observations at discrete time $t_k$, $\mathcal{H}$ the observation operator which maps from the state space to the observation space, and observation error $\eta_k \sim \mathcal{N}(0, R) $. For simplicity, we assume that $R = \sigma_{obs}^2 I_m$ with $\sigma_{obs}^2$ the observation error variance. \\
\ \\
In the analysis step of the EnKF, we incorporate the information from the observations and update the ensemble of forecasts $E_{X_k^b}$.  In the perturbed observation version of the EnKF (see the book by \cite{lewis2006}), the forecast is updated as follows:
\begin{equation} \label{eq:enkf_ana}
    X_k^{a,(i)} = X_k^{b,(i)} + K_X(y_k - \mathcal{H}(X_k^{b,(i)} ) - \tilde{\eta}_k^{(i)} ), \qquad \qquad i = 1, \ldots, N_X,
\end{equation}
where $K_X$ is the Kalman gain matrix and $\tilde{\eta}_k^{(i)}$ is the observation perturbation term, which is assumed to be i.i.d. $\mathcal{N}(0,R)$. The Kalman gain matrix $K_X$ is given by
$$ K_X = \Sigma_{X_k^b, \mathcal{H}(X_k^b) } ( \Sigma_{\mathcal{H}(X_k^b), \mathcal{H}(X_k^b) } + R)^{-1}, $$
where $\Sigma_{X_k^b, \mathcal{H}(X_k^b)}$ is the cross-covariance between the forecast and predicted output, and $\Sigma_{\mathcal{H}(X_k^b), \mathcal{H}(X_k^b) }$ the covariance within the predicted output. The posterior can be characterized using the sample moments of the analysis ensemble:
$$ \mu_{X_k^a} = \frac{1}{N_X} \sum_{i=1}^{N_X} X_k^{a,(i)}, \qquad \qquad \Sigma_{X_k^a} = A_{X_k^a} A_{X_k^a}^T, $$
where $A_{X_k^a}$ contains the analysis ensemble anomalies. In practice, $N_X$ is small and therefore the Monte Carlo estimator $\mu_{X_k^a}$ has a large sampling error, which is proportional to $1/\sqrt{N_X}$. In order to improve the accuracy, we will couple the forecast step \eqref{eq:enkf_for} with a cheaper surrogate $\mathcal{M}_U$ inside a Multi-Fidelity framework.

\subsection{Multi-Fidelity EnKF (MF-EnKF)}

The Multi-Fidelity EnKF (MF-EnKF) is an extension of the standard EnKF algorithm. It was introduced by \cite{popov2021}. The MF-EnKF is based on a Multi-Fidelity approach, see \cite{peherstorfer2018}. The general idea behind this approach is to combine information from a small number of accurate (high-fidelity) estimates with a large number of less accurate (low-fidelity) estimates, in order to reduce the variance and hence the error of the corresponding estimator. \\
\ \\
The original MF-EnKF is derived from a control variate framework. Let $X \in \mathbb{R}^n$ be the random variable representing the full state vector, which we now call the principal variate. We assume that it is expensive to obtain realizations from $X$. Our goal is to estimate the mean of $X$, $\mu_X := \mathbb{E}(X)$. To do this, we introduce two new random variables: a control variate $\widehat{U} \in \mathbb{R}^n$, and an ancillary variate $U \in \mathbb{R}^n$, which should both be cheap to draw samples from. We combine them in the so-called total variate
\begin{equation}
    Z = X - \lambda (\widehat{U} - U  ),
\end{equation}
where $\lambda$ is the gain parameter and $\widehat{U} - U$ the correction term to $X$. The random variable $U$ is assumed to be independent of $X$ and $\widehat{U}$. It is used to estimate the mean of $\widehat{U}$ and to make the estimator unbiased. If the correlation between $X$ and $\widehat{U}$ is large enough, then the variance of $Z$ is reduced compared to $X$. For more details on the control variate framework, see \cite{rubinstein1985}. For simplicity, we assume a scalar $\lambda$, but the multi-fidelity framework can be extended to the general case with matrix weights, see \cite{destouches2024}. \\
\ \\
The idea of the MF-EnKF is now to apply the EnKF to the total variate $Z$ instead of the principal variate $X$. However, we cannot directly sample from $Z$, since $Z$ is not observed, so we use ensembles of principal, control and ancillary variates. Assume that the principal ensemble has size $N_X$, the control ensemble has size $N_{\widehat{U}}$ and the ancillary ensemble has size $N_U$. Typically, $N_U$ is chosen much larger than $N_X$, since it is much cheaper to generate samples from $U$ than samples from $X$. This means that an ensemble for $Z$ is not properly defined, but we can still compute its sample moments based on the 3 separate ensembles. The estimator for the mean of $Z$ is given by
\begin{equation} \label{eq:mf_estimator}
    \mu_{Z} = \frac{1}{N_X} \sum_{i=1}^{N_X} X^{(i) } - \lambda \left( \frac{1}{N_{\widehat{U}}} \sum_{i=1}^{N_{\widehat{U}}} \widehat{U}^{(i) } - \frac{1}{N_U} \sum_{i=1}^{N_U} U^{(i) } \right),
\end{equation}
and the covariance of $Z$ is
\begin{equation} \label{mf_covariance}
    \Sigma_{Z,Z} = \Sigma_{X,X} + \lambda^2 \Sigma_{\hat{U}, \hat{U} } - \lambda \Sigma_{X,\hat{U} } - \lambda \Sigma_{\hat{U}, X} + \lambda^2 \Sigma_{U,U}
\end{equation}
Notice that there is no cross-covariance matrix involving the ancillary variate $U$, since we assume that $U$ is independent of both the principal and control variate. Furthermore, to be able to compute the cross-covariance $\Sigma_{X,\hat{U}}$, we need $X$ and $\hat{U}$ to have the same number of samples, hence we choose $N_{\hat{U}} = N_X$. Similar to the EnKF, we distinguish between the forecast and the analysis step. A visual representation of the MF-EnKF is given in Figure \ref{fig:mf_enkf} below.

\begin{figure}[H]
    \centering 
    \centerline{\includegraphics[scale=0.3]{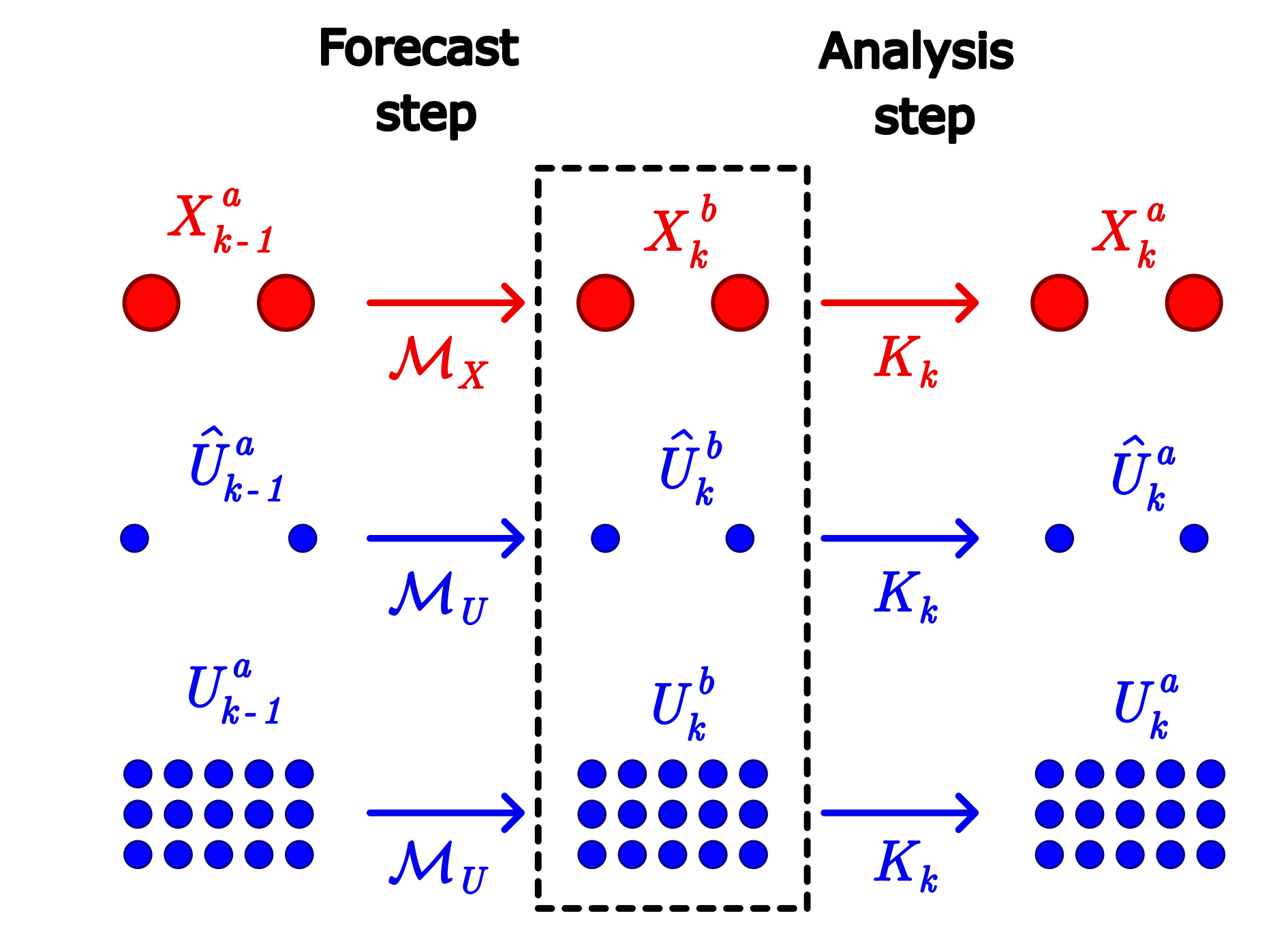}}
    \caption{A visual representation of the MF-EnKF, adapted from Figure 4.1 by \cite{popov2021}. The red ensemble uses the full order model $\mathcal{M}_X$; the blue ensembles use the surrogate model $\mathcal{M}_U$.} 
    \label{fig:mf_enkf}
\end{figure}

\noindent
\textbf{Forecast step of the MF-EnKF} \\
In the forecast step of the MF-EnKF, we assume that we have an ensemble of $N_X$ principal variates, denoted by $\{X_k^{a,(i)} \}_{i=1}^{N_X}$, $N_X$ control variates, denoted by $\{\widehat{U}_k^{a,(i)} \}_{i=1}^{N_X} $, and $N_U$ ancillary variates, denoted by $\{U_k^{(i)} \}_{i=1}^{N_U}$. Similar to the forecast step \eqref{eq:enkf_for} in the standard EnKF, we propagate each of the 3 ensembles forward in time:
\begin{equation} \label{eq: mfenkf_forecast}
\begin{aligned}
    X_{k}^{b, (i) } = \mathcal{M}_X (X_{k-1}^{(i)} ), \qquad  \qquad i = 1, \ldots, N_X, \\
    \widehat{U}_{k}^{b, (i) } = \mathcal{M}_U (\widehat{U}_{k-1}^{(i)} ), \qquad \qquad i = 1, \ldots, N_X, \\
    U_{k}^{b, (i) } = \mathcal{M}_U (U_{k-1}^{(i)} ), \qquad \qquad i = 1, \ldots, N_U,
\end{aligned}
\end{equation}
where $\mathcal{M}_X$ is the expensive forward model and $\mathcal{M}_U$ a cheaper surrogate model. In our case, $\mathcal{M}_U$ is an ML model. The specific choice of ML model depends on the application, see Section \ref{sec:lorenz2005_surr} and \ref{sec:QG_surrogate} for more details. The improved estimate for the mean of $X$ after the forecast step is given by
$$
\begin{aligned}
\mu_{Z_k^b } &= \mu_{X_k^b} - \lambda (\mu_{\widehat{U}^b} - \mu_{U^b} ) \\
&= \frac{1}{N_X} \sum_{i=1}^{N_X} X_{k}^{b,(i) } - \lambda \left( \frac{1}{N_X} \sum_{i=1}^{N_X} \widehat{U}_{k}^{b, (i) } - \frac{1}{N_U} \sum_{i=1}^{N_U} U_{k}^{b, (i) } \right).
\end{aligned}
$$
Notice that we need to run the physical model $N_X$ times and the surrogate model $N_X + N_U$ times in order to produce the forecast. \\
\ \\
\textbf{Analysis step of MF-EnKF} \\
In theory, we can use any variant of the EnKF inside a multi-fidelity framework. We will use the deterministic EnKF \cite{sakov2008}. By using the deterministic EnKF, we avoid the extra tuning process of constructing the proper observation perturbation terms for each of the ensembles, see also the discussion in \cite{popov2021}. In the deterministic EnKF, we decompose the analysis step \eqref{eq:enkf_ana} into an update equation for the means $\mu_{X_k^a}$ and the anomalies $A_{X_k^a}$,
\begin{equation} \label{eq:det_enkf}
  \mu_{X_k^a} = \mu_{X_k^b} + K_X (y_k - \mu_{\mathcal{H}(X_k^b)}),  \qquad \qquad A_{X_k^a} = A_{X_k^b} - \frac{1}{2} K_X A_{\mathcal{H}(X_k^b) }.
\end{equation}
The new analysis ensemble is then calculated as $ X_k^a = \mu_{X_k^a} + \sqrt{N_X - 1} A_{X_k^a} $. We repeat this for each ensemble as follows:
\begin{equation} \label{eq: mf_enkf_ana}
    \begin{aligned}
        X_{k}^a = \mu_X^b + K_Z (y_k - \mu_{\mathcal{H}(X)}^b) + \sqrt{N_X -1} (A_X^b - \frac{1}{2} K_Z A_{\mathcal{H}(X)}^b), \\
        \widehat{U}_k^a = \mu_{\widehat{U} }^b + K_Z (y_k - \mu_{\mathcal{H}(\widehat{U})}^b ) + \sqrt{N_X - 1} (A_{\widehat{U}}^b - \frac{1}{2} K_Z A_{\mathcal{H}(\widehat{U}) }^b ), \\
        U_k^a = \mu_U^b + K_Z(y_k - \mu_{\mathcal{H}(U)}^b) + \sqrt{N_U -1} (A_U^b - \frac{1}{2} K_Z A_{\mathcal{H}(U) }^b ),
    \end{aligned}
\end{equation}
where $K_Z$ is the Kalman gain for the total variate, defined similar to the standard Kalman gain as
\begin{equation} \label{eq:kalman_gain}
    K_Z = \Sigma_{Z_{k}^b, \mathcal{H}(Z_{k}^b) } ( \Sigma_{\mathcal{H}_k(Z_{k}^b), \mathcal{H}(Z_{k}^b) } + R )^{-1},
\end{equation}
where $Z_k^b = X_k^b - \lambda (\widehat{U}_k^b - U_k^b)$ is the background total variate. The covariances that appear in the Kalman gain are calculated as
\begin{equation} \label{eq:cov_mat}
    \begin{aligned}
 \Sigma_{Z, \mathcal{H}(Z)} &= \Sigma_{X,\mathcal{H}(X) } + \lambda^2 \Sigma_{\widehat{U}, \mathcal{H}(\widehat{U} )} - \lambda \Sigma_{X, \mathcal{H}(\widehat{U} ) } - \lambda \Sigma_{\widehat{U}, \mathcal{H} (X) } + \lambda^2 \Sigma_{U, \mathcal{H}(U) }; \\
 \Sigma_{\mathcal{H}(Z), \mathcal{H}(Z)} &= \Sigma_{\mathcal{H}(X),\mathcal{H}(X) } + \lambda^2 \Sigma_{\mathcal{H}(\widehat{U}), \mathcal{H}(\widehat{U} )} - \lambda \Sigma_{\mathcal{H}(X), \mathcal{H}(\widehat{U} ) } - \lambda \Sigma_{\mathcal{H}(\widehat{U}), \mathcal{H} (X) } + \lambda^2 \Sigma_{\mathcal{H}(U), \mathcal{H}(U) }. 
    \end{aligned}
\end{equation}
The cross-covariances we calculate in terms of the ensemble anomalies matrices as $\Sigma_{X_1, X_2} = A_{X_1} A_{X_2}^T$ for ensembles $X_1$ and $X_2$. The cross-covariance is only well-defined for ensembles of the same size, as it is an element-wise comparison. The Kalman gain is shared between the 3 ensembles, so we are not just combining 3 independent EnKF's, but each ensemble uses the information from the other ensembles to produce the analysis. This also means that we only have to compute the Kalman gain once instead of 3 times. A variant on the MF-EnKF, in which each ensemble has a separate Kalman gain matrix, is explored in \cite{hoel2022}. Finally, the total variate analysis is
\begin{equation} \label{eq: tot_var_an_1}
        Z_{k}^a = X_{k}^{a} - \lambda \left(  \widehat{U}_{k}^{a }- U_{k}^{a} \right)
\end{equation}
This shows that we can view the MF-EnKF as 3 parallel EnKF's applied to the principal, control and ancillary variate, where only in the final step we combine them to calculate the total variate.


\section{MF-EnKF with ML surrogate} \label{mf_enkf_ml_surr}

\subsection{Hybrid Modeling}
We propose a version of the MF-EnKF in which the surrogate model $\mathcal{M}_U$ from the forecast step \eqref{eq: mfenkf_forecast} is a ML surrogate, instead of a reduced-order model as used by \cite{popov2021}. This is similar to what is done in hybrid modeling, in which the full model $\mathcal{M}_X$ is often completely replaced by a ML surrogate (\cite{brajard2020}; \cite{bocquet2020}). However, ML surrogates are usually less accurate than the high-resolution model. Hence, by keeping the HR model runs, we hope to gain some accuracy. The forecast step is typically the most expensive part of the EnKF, especially when dealing with large physical models such as weather models. So speeding up the forecast step should lead to the biggest gain in computational efficiency. \\
\ \\
We replace $\mathcal{M}_U$ by a ML surrogate in the following way:
\begin{equation} \label{surr_step}
    U_k^b = \mathcal{M}_{U}(U_{k-1}^a) = U_{k-1}^a + f_{NN} (U_{k-1}^a ; W),
\end{equation}
where $f_{NN}(U_{k-1}; W)$ is a Neural Network (NN) with input $U_{k-1}$ and learnable weights and biases contained in the matrix $W$. In our case, we will consider a Convolutional Neural Network (CNN) and a U-Net architecture, see sections \ref{sec:lorenz2005_surr} and \ref{sec:QG_surrogate} for the details. Notice that in \eqref{surr_step} we fit the Neural Network on the residuals $U_k - U_{k-1}$ and not directly on $U_k$. This is easier to train on, as the steps are smaller and the gradient makes less sudden jumps, see \cite{he2016}.

\subsection{Localization and inflation}
\noindent
\textbf{Localization} \\
Localization is a common method to avoid spurious correlations and increase the rank of the ensemble covariance matrix \cite{houtekamer1998}. The relevant covariances in the MF-EnKF are given by \eqref{eq:cov_mat}, with the Kalman gain matrix calculated as \eqref{eq:kalman_gain}. We apply a localization matrix $\rho$ to the covariances $\Sigma_{Z, \mathcal{H}(Z) }$ and $\Sigma_{\mathcal{H}(Z), \mathcal{H}(Z)}$. For the localization matrix $\rho$ we choose the commonly used Gaspari and Cohn (GC) localization function \cite{gaspari1999}, with a scaling parameter of $c = \sqrt{10/3}$. Note that alternatively we could use separate localization matrices for each of the variates $X, \hat{U}$ and $U$, although this would not guarantee that the combined covariance $\Sigma_{Z,Z}$ is also SPD. \\  
\ \\
\textbf{Inflation} \\
Inflation, as introduced in \cite{anderson1999}, is a way to increase the variability in the ensemble if the ensemble size is small. Since in practice $N_X$ is small, inflation is commonly used to improve the performance of the EnKF. In the MF-EnKF, we will apply an inflation factor to each of the individual analysis anomalies as follows:
$$ A_X^a \rightarrow \alpha_X A_X^a, \qquad A_{\hat{U}}^a \rightarrow \alpha_{\hat{U} } A_{\hat{U} }^a, \qquad A_U^a \rightarrow \alpha_U A_U^a, $$
where $\alpha_X, \alpha_{\hat{U} }$ and $\alpha_U$ are the inflation factors for the principal, control and auxiliary variate. We have to choose $\alpha_{\hat{U} } = \alpha_X$ to maintain the high correlation between $X$ and $\hat{U}$. For simplicity, we take $\alpha_X = \alpha_{\hat{U}} = \alpha_U$. After inflation, we calculate the analysis ensembles.

\subsection{Heuristic ensemble corrections} \label{sec:heuristic}

A key assumption of the Multi-Fidelity EnKF is that $U$ and $\hat{U}$ should have the same mean. This is guaranteed in the limit of infinite ensemble sizes where $N_X$ and $N_U$ go to infinity. However, if $N_X$ is much smaller than $N_U$, the sample mean of $\hat{U}$ can differ from the mean of $U$ due to sampling errors. Therefore, following \cite{popov2021}, we recenter the analysis ensemble means by the sample mean of the total variate $\mu_{Z^a}$:
\begin{equation} \label{recenter}
  \mu_{\hat{U}^a} \rightarrow \mu_{Z^a}, \qquad \mu_{U^a} \rightarrow \mu_{Z^a}.  
\end{equation}
Another important assumption of the Multi-Fidelity EnKF is that $X$ and $\hat{U}$ should be highly correlated. This assumption may not be satisfied if $\mathcal{M}_U$ is not a good enough surrogate model for $\mathcal{M}_X$. \cite{popov2021} achieve this by setting the anomalies matrix $A_{\hat{U}^a}$ for the control variate equal to the projection of the anomalies $A_{X^a}$ onto the reduced space. We ensure the high correlation by explicitly setting them equal:
\begin{equation} \label{ensure_correlation}
    A_{\hat{U}^a} = A_{X^a}.
\end{equation}
This implies that the analysis step for $X^a$ and $\hat{U}^a$ will be exactly the same, but they start from different forecasts $X^b$ and $\hat{U}^b$. This heuristic correction makes sure that the covariance in \eqref{mf_covariance} remains positive definite for any choice of $\lambda$. An alternative approach would be to estimate the multi-fidelity covariance matrix using Log-Euclidean geometry as proposed by \cite{maurais2023}.


\section{Application to Lorenz-2005 model} \label{sec:lorenz2005}

\subsection{The Lorenz-2005 model}
We will first test our MF-EnKF algorithm on the Lorenz-2005 equations \cite{lorenz2005}, which have also been used as a test case in \cite{rainwater2013} for the Mixed-Resolution EnKF. The so-called model II from \cite{lorenz2005} is a generalization of the more well-known Lorenz-96 model \cite{lorenz96}, which is one of the most common benchmark models for data assimilation. The Lorenz-96 equations are given by
\begin{equation}\label{lorenz96}
    \frac{d X_i}{dt} = (X_{i+1} - X_{i-2}) X_{i-1} - X_i + F , \qquad \text{ for } i = 0, \ldots, n-1,
\end{equation}
with periodic boundary conditions $X_n = X_0, X_{-1} = X_{n-1} $ and $X_{-2} = X_{n-2}$. The equations consist of an advection term $(X_{i+1} - X_{i-2})X_{i-1}$, dissipation term $X_i$ and external forcing term $F$. For certain choices of $F$, the model is chaotic and can be viewed as a simplified atmospheric model, which describes some atmospheric quantity on a circle located at constant latitude, divided into $n$ equally spaced grid points. \\
\ \\
The original Lorenz-96 equations \eqref{lorenz96} do not allow for spatial interpolation, since increasing the spatial resolution results in more copies of the same equation, but not in a finer mesh. On the other hand, the Lorenz-2005 equations include a mesh size that can be adjusted to the dimension. The spatial interpolation will be needed to generate lower-resolution surrogates as an alternative to the ML surrogate. The Lorenz-2005 equations are given by
\begin{equation} \label{lorenz2005}
    \frac{d X_i}{dt} = [ X,X ]_{K,i} - X_i + F,
\end{equation}
where 
$$ [X,X]_{K,i} := \sum_{j=-J}^{J} \textcolor{white}{}^\prime \ \sum_{k=-J}^{J} \textcolor{white}{}^\prime \ \frac{1}{K^2} (-X_{i-2K-k} X_{i-K-j} + X_{i-K+j-k} X_{i+K+k} ), $$
with $K$ an even integer, $J = K/2$ and $\sum \textcolor{white}{}^\prime$ equal to the normal summation, with the first and last term divided by two to preserve the symmetry. That is,
$$ \sum_{k=0}^n \textcolor{white}{}^\prime \  a_k := \frac{a_0}{2} + \frac{a_n}{2} + \sum_{k=1}^{n-1} a_k. $$
The parameter $K$ is a smoothing parameter that determines the length over which we take the average. A larger $K$ results in a smoother signal. Similar to \cite{lorenz2005}, we choose $K=32$, $n=960$ and $F = 15$, which makes the system chaotic. If $K$ is odd, we define $J = (K-1)/2$ and $\sum \textcolor{white}{}^\prime$ reduces to the normal summation $\Sigma$. Notice that for $K=1$, Equation \eqref{lorenz2005} reduces to the standard Lorenz-96 model. For the implementation, we use the Lorenz-2005 model from DAPPER \cite{raanes2024}, a data assimilation library in Python. 

\subsection{The Lorenz-2005 ML surrogate} \label{sec:lorenz2005_surr}

For the ML surrogate model, we use a Convolutional Neural Network (CNN) \cite{goodfellow2016}. We choose the architecture of the CNN in such a way that we stay close to the original equations, similar to \cite{brajard2020}, where a CNN was created to emulate the Lorenz-96 equations. We reformulate \eqref{lorenz2005} as
$$ [X,X]_{K,i} = - W_{i-2K} W_{i-K} + \sum_{j=-J}^J \textcolor{white}{}^\prime \ \frac{1}{K} W_{i-K+j} X_{i+K+j}, \qquad  W_i := \sum_{k=-J}^J \textcolor{white}{}^\prime \ \frac{1}{K} X_{i-k}. $$
We can view $W_i$ as the $i$-th element of the convolution of the kernel $[ \frac{1}{K} \frac{1}{K} \cdots \frac{1}{K} ] \in \mathbb{R}^{K+1}$ with the signal $X \in \mathbb{R}^n$. Similarly, we can see that, if we rewrite $W_{i-2K} = \sum_{j=-J}^J \textcolor{white}{}^\prime\ \frac{1}{K} X_{i-2K-j} = \sum_{j=2K-J}^{2K+J} \frac{1}{K} X_{i-j}$, then $W_{i-2K}$ is the $i$-th element of the convolution of the kernel $[ \frac{1}{K} \cdots \frac{1}{K} ] \in \mathbb{R}^{2K+1} $ with the signal $X$. In this way, we can replace each $W_i$ with a convolutional layer and represent $[X,X]_{K,i}$ as
$$ \text{ConvLayer}(5K) \cdot \text{ConvLayer}(3K) + \text{ConvLayer}(4K) \cdot \text{ConvLayer}(3K), $$
where ConvLayer$(k)$ denotes a Convolutional Layer with kernel size $k$. In total the network has 89,699 parameters. The input is first put through a batch-normalization layer. The output layer has 1 filter of kernel-size 1, so that the output dimension is again 960 by 1. Each convolutional layer uses a ReLU activation function, except for the last layer where a linear activation function is used. Furthermore, periodic padding is added to each layer, since the boundary conditions are periodic. The architecture is implemented in PyTorch, and training and testing is done on a NVIDIA RTX A5000 GPU. For an overview of the full Neural Network architecture, see Figure \ref{fig:cnn_architecture} below.

\begin{figure}[H]
    \centering
    \includegraphics[scale=0.5]{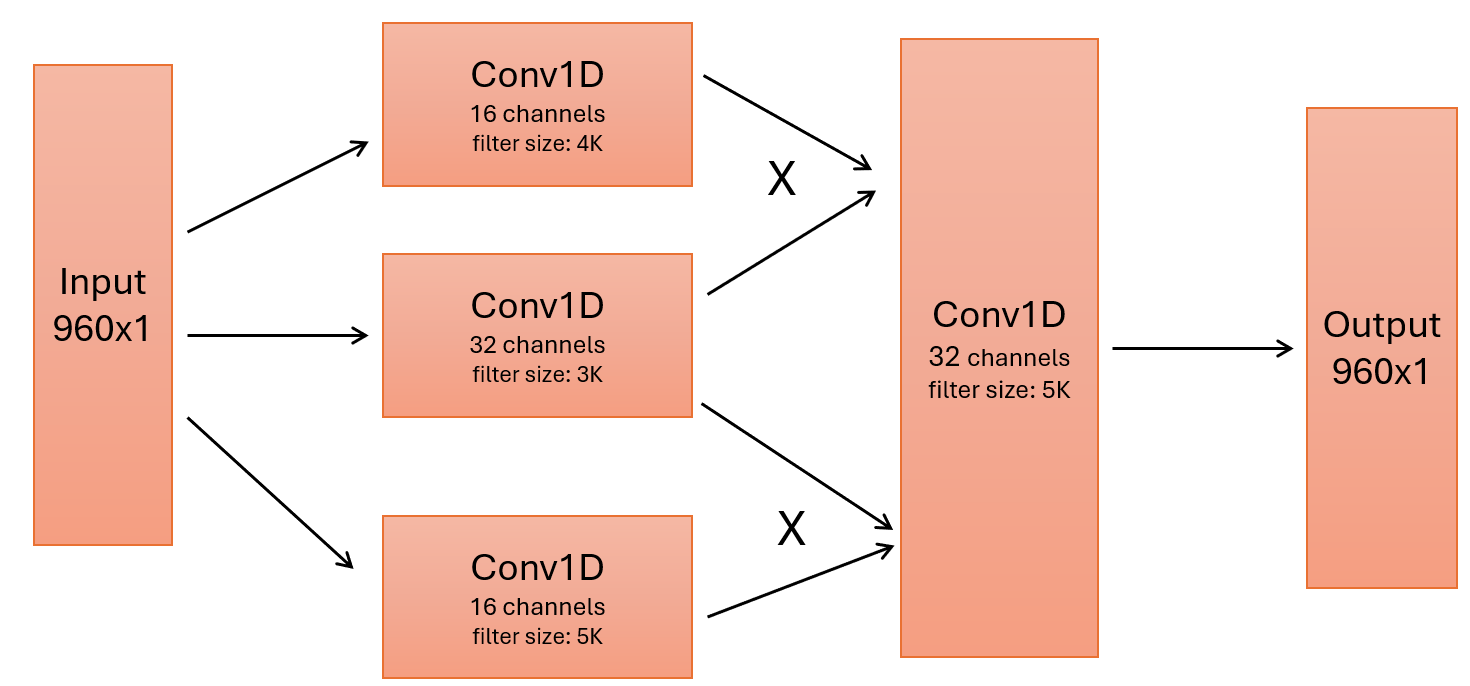}
    \caption{CNN architecture. The input goes through 3 Conv1D blocks. The output of these blocks is then multiplied together and put through another convolutional layer. The batch-normalization and output layer are not shown in the picture.}
    \label{fig:cnn_architecture}
\end{figure}

\noindent
The data for training the CNN is generated from direct model simulations of \eqref{lorenz2005}, with an initial spin-up of 2 years. After the spin-up, the model is run for 5 years to create the training data set. Test data is created by running the model for 2 more years. The batch size is tuned through a grid search with values in $\{32,64,128,256,512\}$. We first train the network with a learning rate of $10^{-3}$ for 100 epochs and then another 100 epochs with learning rate $10^{-4}$. The lowest test mean squared error (MSE) loss was achieved for a batch size of 64.

\subsection{Experimental setup}
For the setup of the experiments, we use the same settings as in \cite{lorenz2005}. We discretize \eqref{lorenz2005} in time using the RK4 method with step size $\Delta t = 0.025$, which corresponds to about 3 hours. We start with an initial condition $x_0 \in \mathbb{R}^n$ whose elements are sampled independently from the $\text{Unif}(0,1)$ distribution. In order to minimize the influence of the initial condition, we spin up the model for 2 years. We assume observations are available every $\Delta t = 0.05$ time steps (corresponding to 6 hours) and at $m=40$ equidistant locations in space, corresponding to about $4\%$ of the total state. We assume a noise level of $\sigma_{obs} = 2.0$ (signal-to-noise ratio equal to $27\%$). \\
\ \\
On the data assimilation side, we create the initial ensemble of the EnKF by adding to the true state Gaussian noise with mean 0 and standard deviation 5 (about the size of the true variability of the signal). Data assimilation is performed for $T=1000$ time steps (about 4 months), with a burn-in of 100. Since sample sizes are kept small to mimic the real-life case where computational power is limited, the ensembles have a large sampling error. In order to minimize the influence of this error, we generate new observations with $M=10$ different seeds and then average the results. The EnKF is tuned  with an optimal inflation factor and localization, where we use a periodic $l_1$ distance inside the Gaspari-Cohn function. The MF-EnKF has a fixed tuning parameter of $\lambda = 0.5$. The performance is measured by calculating the average root mean squared error (RMSE) over the $M=10$ observation trajectories and $T=1000$ time steps.

\subsection{Results}

\subsubsection{Comparison with EnKF} \label{lorenz05_comp_enkf}

We compare our version of the MF-EnKF for different values of $N_X$ and $N_U$ with an optimally tuned EnKF that uses $N_X = 100$ HR ensemble members. Furthermore, we compare to a 'baseline' approach, which consists of merging the HR ensemble with the ML surrogate ensemble. In the forecast step these ensembles are propagated separately, but in the analysis step we treat them as one ensemble. This baseline serves as the naive approach in which the difference between the HR and ML model is not taken into account. For the MF-EnKF, we apply no localization and only apply inflation of $\alpha = 1.02$ for $N_X = 2$. Results are given in Figure \ref{lorenz_2005_main_res} below.

\begin{figure}[H]
    \centering
    \includegraphics[scale=0.7]{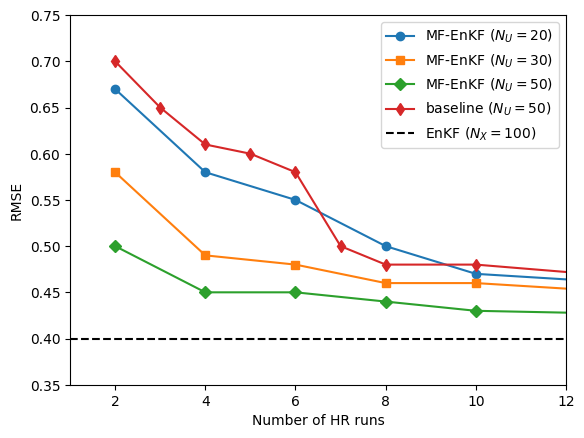}
    \caption{Performance of MF-EnKF against an optimally tuned EnKF for different number of full runs $N_X$. The solid lines indicate the MF-EnKF and the baseline with a changing number of high-res runs. The dashed line is the EnKF RMSE when using $N_X = 100$.}
    \label{lorenz_2005_main_res}
\end{figure}
\noindent
In general, we can see that the performance of the MF-EnKF gets close to that of the EnKF with an ensemble size of 100, but at a lower computational cost. The largest benefit of adding surrogate runs $N_U$ is when the number of full runs $N_X$ is small, since the EnKF is not stable for such a small number of the ensemble size. Furthermore, the MF-EnKF outperforms the baseline, indicating that weighting the ensembles is beneficial.

\subsubsection{Comparison with LR surrogate}
Instead of the ML surrogate, the original MF-EnKF paper \cite{popov2021} uses a lower-resolution surrogate model created using a proper orthogonal decomposition (POD). Therefore, it is interesting to compare the performance of the ML surrogate with different low-resolution models. We generated these from the Lorenz-2005 equations \eqref{lorenz2005} with a fixed ratio $\frac{r}{K} = 30$, where $r$ is the dimension of the low-resolution model. We create 3 low-resolution models, denoted by $m120$, $m240$ and $m480$, which use dimensions $r = 120$, $r = 240$ and $r = 480$ respectively. Each low-resolution model samples $r$ equidistant points from the state vector $X \in \mathbb{R}^n$ and propagates these points forward in time. After that, linear interpolation is applied to map the points back to the fine grid of dimension $n=960$. \\
\ \\
To compare the quality of the different surrogates, we compute their forecast RMSE. These forecasts are generated following the same procedure as \cite{lorenz2005}. We sample a random initial condition $x_0 \sim \text{Unif}(0,1)$. Then the model is propagated forward in time for 120 days. The state at $T=120$ days is then used as the initial state for the forecasting experiments. This process is repeated $M=100$ times. The resulting average forecast RMSE for each surrogate is given in Table (\ref{lowres_forecast}) below. For short lead times, the performance of the NN surrogate is between $m240$ and $m480$. For longer lead times, the NN becomes less stable as expected.

\begin{table}[H]
    \centering
      \caption{Prediction accuracy for the low-resolution surrogates of dimension $r=120$ (m120), $r=240$ (m240) and $r=480$ (m480), and the ML surrogate (denoted by NN) at different lead times. In bold the minimum RMSE is highlighted.}
    \begin{tabular}{c|c|c|c|c}
       \textbf{Lead time}  & \textbf{m120} & \textbf{m240} & \textbf{m480} & \textbf{NN} \\ 
       \hline
        6h & 0.34 & 0.089 & \textbf{0.022} & 0.042 \\
        1 day & 0.41 & 0.10 & \textbf{0.024} & 0.11 \\
        1 week & 2.83 & 0.93 & \textbf{0.21} & 1.37 \\
    \end{tabular}
    \label{lowres_forecast}
\end{table}

\begin{figure}[H]
    \centering
    \includegraphics[scale=0.7]{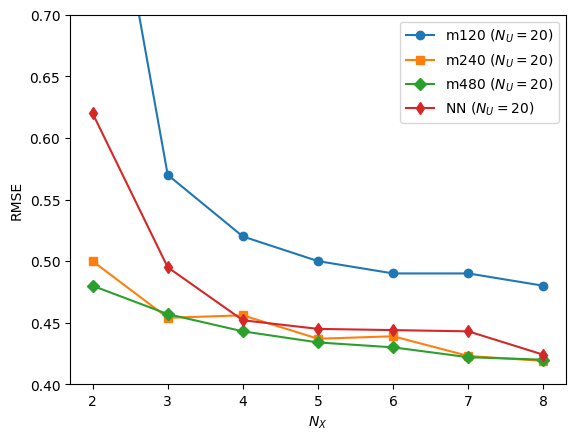}
    \caption{RMSE values for MF-EnKF with the Neural Network (NN) surrogate versus a MF-EnKF with a low-resolution surrogate. The number of surrogate runs is fixed at $N_U = 20$.}
    \label{fig:NN_vs_lowres}
\end{figure}

For each of the surrogate models, we run the MF-EnKF with a fixed number of surrogate runs $N_U$ and a different number of full runs $N_X$. We fix $N_U = 20$ since for this value we can see bigger changes in performance when we change $N_X$. For the LR model $m120$ we use inflation $\alpha = 1.04$, for LR $m240$, $m480$ and the NN we use inflation $\alpha = 1.02$. For the NN, we apply the recentering of the analysis means for stability as in \eqref{recenter}.  We compare the RMSE values in Figure (\ref{fig:NN_vs_lowres}). We can see that the NN surrogate has similar performance to the low-resolution surrogates, but with higher RMSE for smaller $N_X$ values, indicating the instability of the NN. In general, the performance of each of the surrogate models is in line with their 1-step ahead forecast accuracy. This suggests that performance of the MF-EnKF improves if a better surrogate model is chosen, at least for this test case. In practice, the performance difference between the NN and the LR models is expected to be even bigger, as decreasing the resolution can result in missing important grid-specific features, which can heavily influence the predictions.

\subsubsection{Computational cost} \label{sec:comp_cost_lorenz}

\begin{figure}[H]
    \centering
    \includegraphics[scale=0.7]{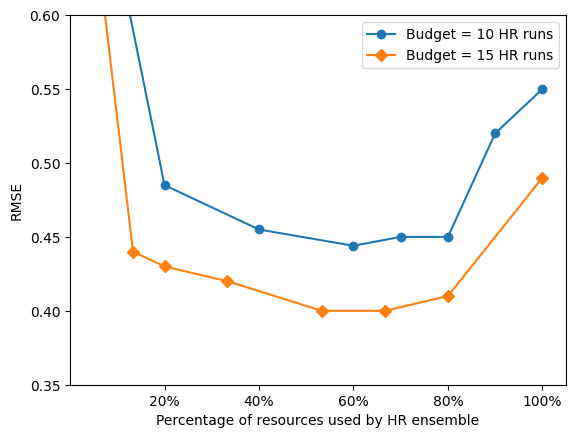}
    \caption{RMSE for the MF-EnKF for different pairs $(N_X,N_U)$, expressed in percentage of resources used by HR ensemble. For a budget of 10 HR runs, a percentage of $100 \%$ corresponds to $(N_X=10,N_U=0)$ and a percentage of $0 \%$ corresponds to $(N_X=0,N_U=100)$. Minimum RMSE is attained at about $50 \%$ in both cases.}
    \label{fig:comp_cost}
\end{figure}

The comparison made in Figure \ref{lorenz_2005_main_res} is not entirely fair, since the MF-EnKF can use the information of extra surrogate runs and so effectively has a larger ensemble size than the EnKF. In order to make a fairer comparison, we assume a fixed computational budget. In practice, it is expected that the ML model will be much faster than the computationally heavy full model, see also Table \ref{tab:ml_weather} from the introduction. In our small test model, computational times will depend a lot on our specific implementation, and the fact that the ML model can use the GPU for its computations. Therefore, we make the conservative assumption that our ML model is 10 times faster than the physical model. Furthermore, we assume a fictional computational budget of 10 physical model runs. Within these constraints, we can make an optimal choice of assigning computational power to the different ensembles. Similar to Figure 8 from \cite{barthelemy2024}, we make a plot of different combinations of high and low fidelity runs, within the fixed budget of $N_X = 10$ full runs. Figure \ref{fig:comp_cost} displays the performance of the MF-EnKF for different $(N_X, N_U)$ pairs.

\noindent
We can see that adding surrogate runs in exchange for fewer full runs results in a better performance, with a minimum RMSE of 0.44 attained at the optimal combination of $N_X = 5$ full runs and $N_U = 50$ surrogate runs, corresponding to a $50 \%$ split between HR resources and ML resources. In general, there seems to be a large region of pairs $(N_X, N_U)$ for which the RMSE is quite stable. However, if we remove too many full runs, performance starts to deteriorate again. This agrees with what we would expect: not completely throwing away the information from the original HR model has a stabilizing effect on the performance compared to only using the ML surrogate.

\subsubsection{Influence of $\lambda$ parameter}

\begin{figure}[H]
    \centering
    \includegraphics[scale=0.5]{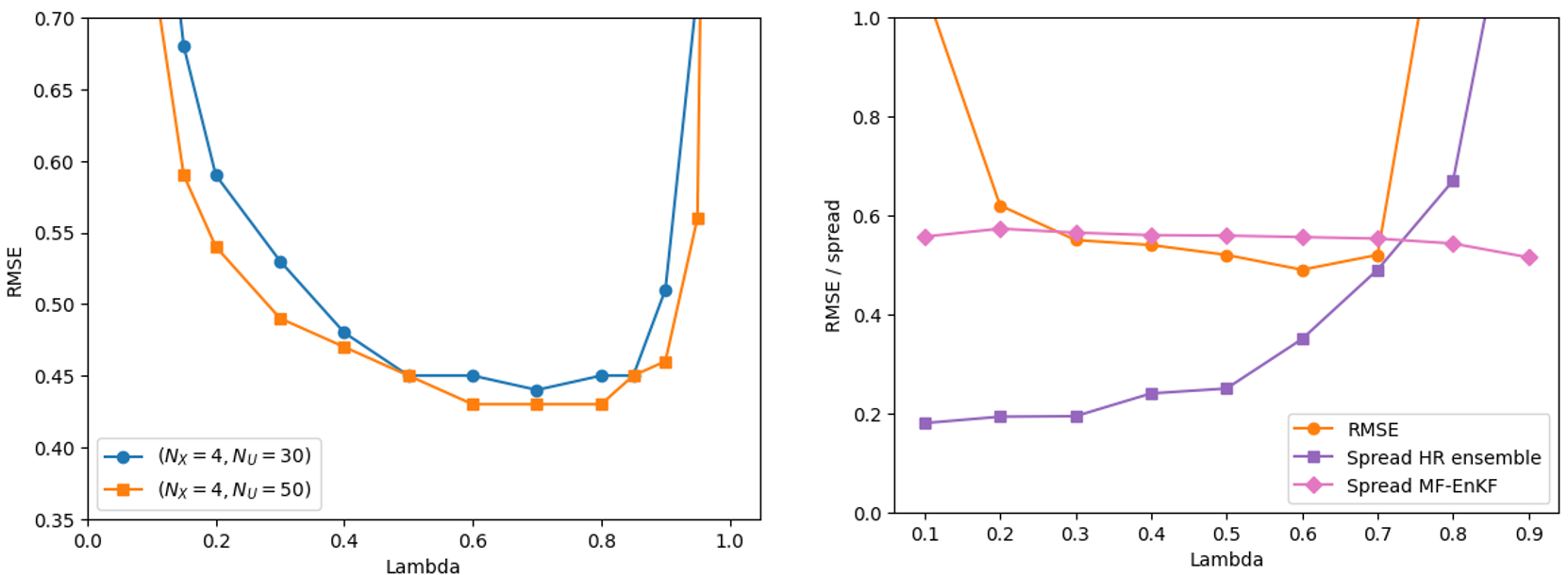}
    \caption{Left: influence of the hyper-parameter $\lambda$ on the RMSE for the combinations $(N_X,N_U) = (4,30)$ and $(4,50)$. Right: spread of the HR and MF-EnKF ensemble against the RMSE error against different $\lambda$ values, for fixed combination $(N_X = 4, N_U = 20)$.}
    \label{fig:infl_of_lambda}
\end{figure}

An extra hyper-parameter, $\lambda$, is introduced when using the MF-EnKF. We would like to know how much influence this parameter has and how we can choose its value. Until now, we have been using a fixed value of $\lambda = 0.5$. For certain choices of $\lambda$, the results are unstable and we add an inflation of $\alpha = 1.01$ to each ensemble to stabilize the results. In Figure \ref{fig:infl_of_lambda} we show the influence of $\lambda$ for a fixed combination of full and surrogate runs. We can see very similar behavior as for the case where we assumed a fixed computational budget. Furthermore, if we pick $\lambda \in [0.4, 0.9]$, then the RMSE is more or less constant, implying that $\lambda$ does not require extensive tuning. This is in agreement with Figure 2 from \cite{rainwater2013}. Furthermore, the chosen value of $\lambda = 0.5$ we used before in the experiments seems to be close to optimal. Besides the influence of $\lambda$ on the RMSE, we also show the influence on the spread of the ensemble. We can see that the MF-EnKF provides an improved estimate of the true error compared to the spread of only the HR ensemble. This holds again for a large range of $\lambda$ values.

\section{Application to QG model} \label{qgmodel}

\subsection{The QG model}
In this section we will look at the Quasi-Geostrophic (QG) model, see \cite{pedlosky1987}, which is a 2-dimensional model and has a state dimension of $128^2 = 16384$. This model has also been used as a test model in the context of the Multi-Fidelity Ensemble Kalman Filter and other EnKF variants (\cite{sakov2008}; \cite{popov2021}; \cite{barthelemy2022}; \cite{barthelemy2024}). \\ 
\ \\
The standard DA test setting for the QG model was introduced by \cite{sakov2008}. They use an implementation in Fortran, which is included in the data assimilation library DAPPER \cite{raanes2024} in Python. In this paper, we use the implementation by \cite{thiry2022}, which is implemented in PyTorch and allows us to compare the computation times with the ML model in a more fair manner. The disadvantage is that this version has not yet been used to test DA approaches. To make it comparable to other papers, we will rewrite the parameters to match those of the standard DA test setting from \cite{sakov2008}. \\
\ \\
The QG model version by \cite{thiry2022} describes the pressure $p$ and potential vorticity $q$ in different layers in the ocean, on a rectangular domain. In the standard setting, a reduced-gravity model is used, which is equivalent to a 2-layer QG model with infinite depth for the second layer. In terms of the streamfunction $\psi$, the QG equation is given by:
\begin{equation}
    \partial_t \left( \Delta \psi - \frac{f_0^2}{H g'} \psi \right) = J(\Delta \psi, \psi) - {\beta \psi_x} + {F_0 \sin \left( \frac{ 2 \pi y }{L} \right)} + {a_2 \Delta^2 \psi} - {a_4 \Delta^3 \psi},
\end{equation}
where $q = \Delta \psi - \frac{f_0^2}{H g'} \psi$ is the potential vorticity, $J(a,b) = a_x b_y - a_y b_x$ the Jacobi operator, $\beta \psi_x$ the Coriolis gradient term, $F_0 \sin \left( \frac{2 \pi y}{L} \right)$ the wind forcing, $a_2 \Delta^2 \psi$ the horizontal friction and $a_4 \Delta^3 \psi$ the biharmonic friction. The parameter values are chosen as to match the setting from \cite{sakov2008}. For the derivation of the parameters, we refer to Appendix \ref{app:QG_param}. An overview is presented in Table \ref{tab:QGparam} below. 

\begin{figure}[H]
    \centering
    \includegraphics[scale=0.7]{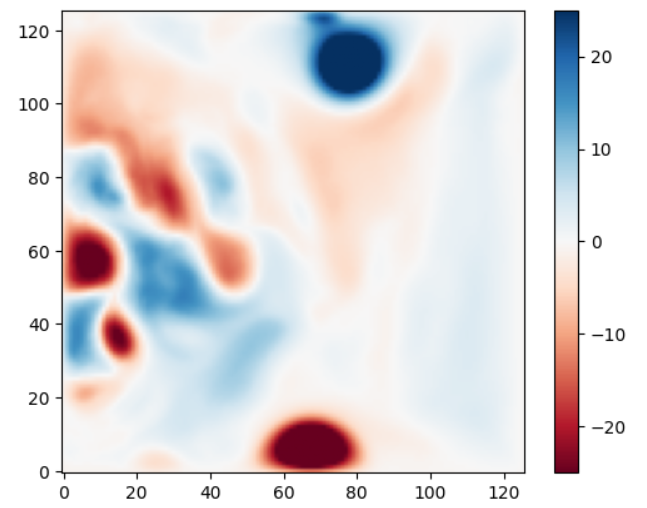}
    \caption{Snapshot of the streamfunction $\psi$ for the chosen parameter set. Darker colors indicate higher streamfunction intensity.}
    \label{QG_snapshot}
\end{figure}

\begin{table}[H]
    \centering
    \caption{Parameters of the QG model}
    \begin{tabular}{c|c|c}
    \textbf{Parameter} & \textbf{Meaning} & \textbf{Value} \\
    \hline
       $f_0$  & Mean coriolis & $8.4 \cdot 10^{-5} (s^{-1}) $  \\
        $\beta $ & Coriolis gradient & $1.88 \cdot 10^{-11} (m^{-1} s^{-1}) $ \\
        $\theta$ & Latitude & 35 (degrees) \\
        $\rho_0$ & Mean ocean density & $1000$ ($kg/m^3$) \\
        $L$ & Domain size & $3072$ ($km$) \\
        $g'$ & Reduced gravity & 0.025 ($m/s^{2}$) \\
        $H$ & Upper layer depth & 1664 ($m$) \\
        $\tau_0$ & Wind stress magnitude & 1.7 $\cdot 10^{-4}$ ($m^{2} s^{-2}$) \\
        $a_2$ & Horizontal friction coefficient & $0 \ (m^2 s^{-1}$) \\
        $a_4$ & Biharmonic friction coefficient & $1 \cdot 10^{10} \ (m^4 s^{-1} ) $ \\
        $\Delta t$ & Time resolution & 6 ($h$) \\
        $n_x, n_y$ & Grid size & 128 \\
        $\Delta x, \Delta y$ & Spatial resolution & 24 \ ($km$) \\
    \end{tabular}
    \label{tab:QGparam}
\end{table}
\noindent
Furthermore, we use a free slip boundary condition, and 0 initial condition $(\psi_0 = q_0 = 0$). For the spatial discretization, an Arakawa scheme is used \cite{arakawa1981} and finite differences for the Helmholtz equation for $q$. For more details on the numerical scheme, see \cite{hogg2003}. For the time discretization, we use the RK4 method with a time step of $\Delta t = 6$ hours. A snapshot of the streamfunction $\psi$ using the settings of the QG model as described above can be seen in Figure \ref{QG_snapshot}. One can clearly see the westward intensification, which is typical for these models.

\subsection{The QG model ML surrogate}
\label{sec:QG_surrogate}

\begin{figure}[H]
    \centering
    \includegraphics[scale = 0.4]{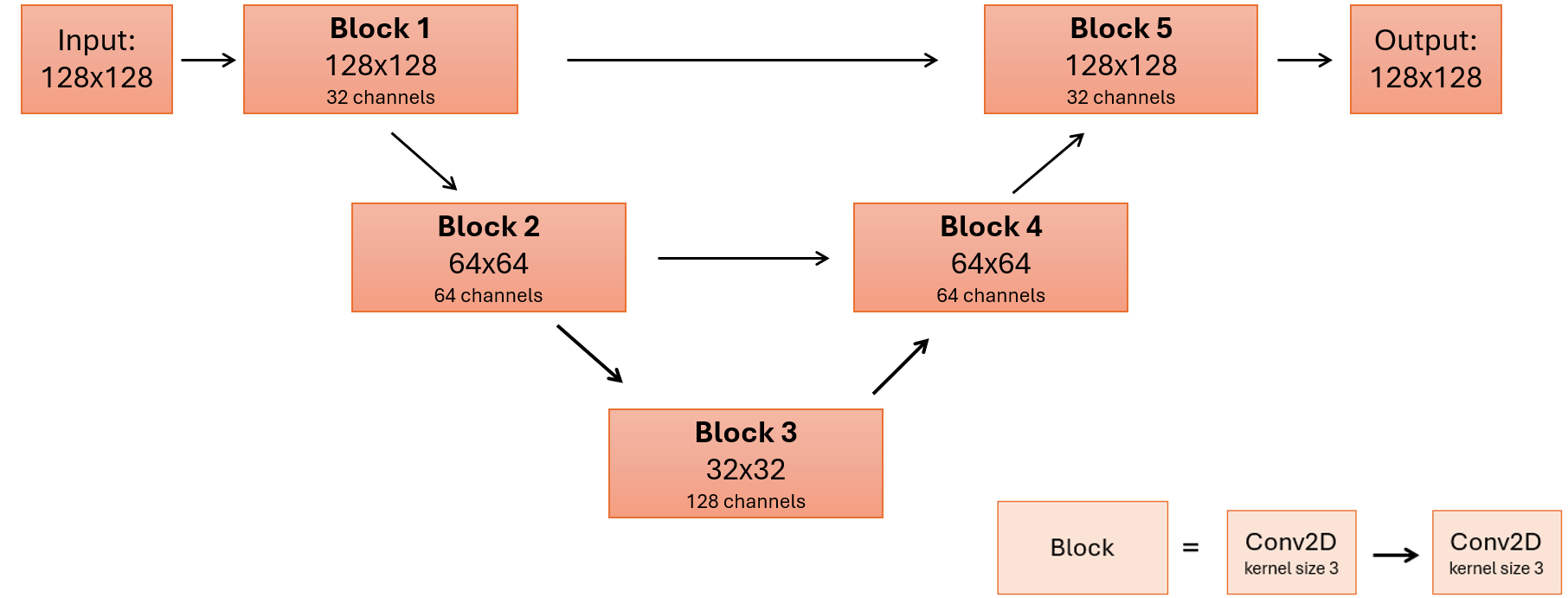}
    \caption{U-Net architecture. Long horizontal arrows represent skip connections, downward pointing arrows represent the down-sampling and upward pointing arrows represent up-sampling. Each block consists of 2 convolutional layers, as given in the lower-right corner.}
    \label{fig:unet_architecture}
\end{figure}

For the QG model we use a U-Net architecture \cite{ronneberger2015}, which was also chosen by \cite{lu2025} for a similar QG model. The U-Net architecture consist of several convolutional blocks, containing a convolutional layer followed by a down-sampling (through max-pooling) or up-sampling (through up-convolution) layer. In each down-sampling step the image resolution is decreased by a factor of 2, whereas the number of channels is increased by a factor 2. In the up-sampling step this is the other way around. Furthermore, skip-connections are added between layers of the same resolution. We use ReLU activation functions in each convolutional layer, and a linear activation function for the output layer. The total number of parameters is 465,953. See Figure \ref{fig:unet_architecture} for a visual overview of our U-Net architecture.

\noindent
The training data for the U-Net is generated by choosing a random initial condition from 2000 snapshots of a long run of length $5 \cdot 10^5$ (corresponding to about 340 years), and integrating forward the initial condition for $20,000$ time steps (10 years). Test data is generated from a different random initial condition and integrated for $1000$ time steps (2 years). Similar as in Section \ref{sec:lorenz2005_surr}, the ML surrogate is trained to predict the increments rather than the full forward operator. We train the ML surrogate to immediately predict 4 time steps ahead, as this is also the observation frequency we will be using later. We use a learning rate of 0.001. The batch size is determined through a grid search with values in $\{8, 16, 32, 64, 128, 256\}$. The number of epochs is determined based on visual inspection of the test RMSE. We find an optimal batch size of 16. The total training time was 30 minutes on GPU.

\subsection{Experimental setup}
For the setup of the numerical experiments, we mostly follow the setting of \cite{sakov2008}. We take a spatial resolution of $128$ by $128$ and a time resolution of $\Delta t = 6$ hours. The true streamfunction is generated by choosing an initial condition at random from the snapshots generated before and propagating forward for $K_{da} = 1500$ time steps. The initial ensembles are generated by choosing $N_X$ (or $N_U$) other initial conditions from these snapshots at random. Inside the MF-EnKF, we choose the deterministic EnKF variant, with no inflation and the same optimal localization radius of 5 as in \cite{sakov2008} and Euclidean distances for the Gaspari-Cohn function. We perform analysis every 4 time steps (equal to 1 day). The tuning parameter $\lambda$ is set to $0.5$. For the observations, we take 344 diagonal measurements of the streamfunction $\psi$ at an angle of 66 degrees, mimicking the behaviour of satellite tracks. The observations are corrupted with Gaussian noise with a standard deviation of $\sigma_{obs} = 2.0$. Results are averaged over $M=10$ different observation realizations. Initially, the experimental results showed some unstable behavior at the boundary of the domain. Therefore, we choose to only assimilate the interior of the domain, which is similar to what is done in \cite{sakov2008}.

\subsection{Results}

\subsubsection{Comparison with EnKF}

We apply the MF-EnKF to the QG model and test the influence of the $N_X$ and $N_U$ values. We take $N_X \in \{1,2,\ldots,10 \}$, which mimics the small values in real-life, and for $N_U$ we choose values in $\{25, 50, 100\}$. We compare with an optimally tuned EnKF with an ensemble size of $N_X = 100$. Furthermore, we compare to the same baseline approach as before in Section \ref{lorenz05_comp_enkf}. The results can be seen in Figure \ref{fig:MFEnKF_vs_EnKF_QG} below. In general, we see that adding more HR runs or adding more ML runs improves the performance, as expected. The RMSE values obtained are similar to those in \cite{sakov2008} and \cite{barthelemy2024}. Furthermore, the MF-EnKF can achieve a performance similar to that of the EnKF with 100 HR runs, but with reduced computational cost. Moreover, the MF-EnKF outperforms the naive baseline approach, even for a lower number of surrogate runs. This suggests that weighting the ensembles is better than naively putting them together.

\begin{figure}[H]
    \centering
    \includegraphics[scale=0.7]{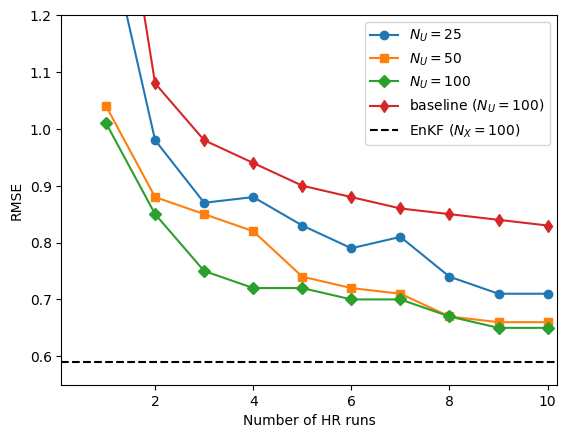}
    \caption{Performance in terms of RMSE of MF-EnKF versus standard EnKF and baseline approach for different $N_X$ values.}
    \label{fig:MFEnKF_vs_EnKF_QG}
\end{figure}

\subsubsection{Comparison with LR surrogate}
As an alternative to the ML surrogate, we can use a LR surrogate. The quality of the LR surrogate will depend on the low-resolution grid size and the Rossby radius of our QG model. For our parameter setting, the Baroclinic Rossby radius is equal to
$$ R_d = \frac{ \sqrt{H g'} }{f_0} \approx 76 \ \text{km}. $$
Although this is an unrealistic value (see \cite{chelton1998} for realistic values), one can see from this value that the resolution of $n_x = 128$ is eddy-resolving, since $\Delta x = \frac{3072 \text{ km}}{128} = 24 \text{ km} \leq \frac{1}{2} R_d = 38 \text{ km} $. The resolution $n_x = 64$ (so $\Delta x = 48$ km) is eddy-permitting, but not eddy-resolving. The resolution $n_x = 32$ (so $\Delta x = 96$ km) is neither. \\
\ \\
First, we compare the forecast accuracy of the LR models with the ML model. We choose 100 initial conditions sampled from the truth run and compute forecasts for each one. The final forecast RMSE is then the average over those values. The results for different lead times are given in Table \ref{tab:qg_forecast}.

\begin{table}[H]
    \centering
    \caption{Forecast RMSE for NN and LR models with resolutions 64x64 (m64) and 32x32 (m32). Minimum RMSE values are highlighted in bold.}
    \begin{tabular}{c|c|c|c}
       \textbf{Lead time} & \textbf{m32} & \textbf{m64} & \textbf{NN} \\
       \hline
        6h & 0.82 & 0.27 & - \\
        24h & 1.53 & 0.68 & \textbf{0.22} \\
        1 week & 4.65 & 2.63 & \textbf{0.54} \\
    \end{tabular}
    \label{tab:qg_forecast}
\end{table}

\begin{figure}[H]
    \centering
    \includegraphics[scale=0.7]{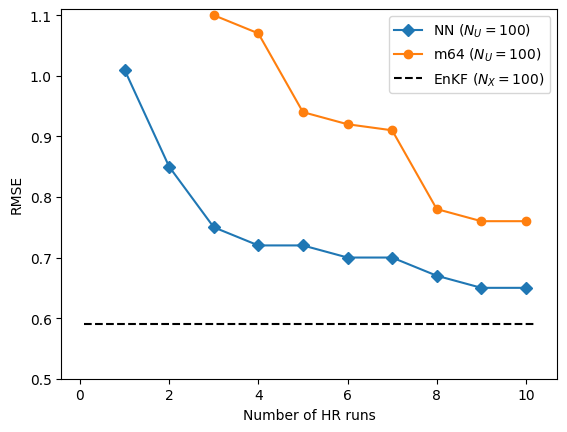}
    \caption{RMSE performance for the ML model and m64 model for different number of HR runs $N_X$.}
    \label{fig:rmse_ML_vs_LR}
\end{figure}

\noindent
Note that the NN is trained to predict 24h ahead, so it needs fewer time steps to roll out the forecast. This is also why the NN remains more stable for longer lead times than previously in Table \ref{lowres_forecast}. Since both the ML model and the original model are implemented on the GPU, we can now more fairly compare the computational times. The average 24h forecast time for the ML model is about 10 times faster. In Figure \ref{fig:rmse_ML_vs_LR} we display the MF-EnKF $(\lambda=0.5)$ RMSE results for the ML surrogate versus the m64 surrogate, with a fixed number of $N_U = 100$ surrogate runs. We can see that the performance of the surrogates is in line with their forecast accuracy, similar to what we observed in Figure \ref{fig:NN_vs_lowres}.

\subsubsection{Computational cost}

\begin{figure}[H]
    \centering
    \includegraphics[scale=0.7]{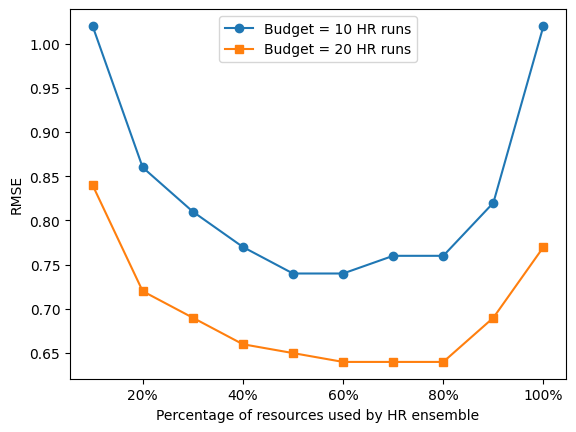}
    \caption{RMSE for different $(N_X, N_U)$ pairs, for total assumed budget of $N_X = 10$ or $N_X = 20$ HR runs. The percentage of HR resources used is $N_X$ divided by the total budget. The remaining budget is used for the $N_U$ ML runs.}
    \label{fig:fixed_comp_budget_QG}
\end{figure}

The computational cost is the biggest limiting factor in practice. Similar to the Lorenz-2005 case, we will assume that the ML model is 10 times faster than the QG model, with a total budget of 10 or 20 HR runs. There are then several ways to divide into $N_X$ and $N_U$ runs within this budget. In Figure \ref{fig:fixed_comp_budget_QG} we show the performance of the MF-EnKF for different combinations $(N_X, N_U)$, in terms of the percentage of resources used for the HR ensemble. We can see that the best performance is a combination of HR and ML runs. In general, there is a large region of $(N_X, N_U)$ pairs for which the RMSE is close together. This is similar to what we observed in Section \ref{sec:comp_cost_lorenz} and what was observed in Figure 5 in \cite{barthelemy2024}.

\subsubsection{Influence of $\lambda$ parameter}

\begin{figure}[H]
    \centering
    \includegraphics[scale=0.45]{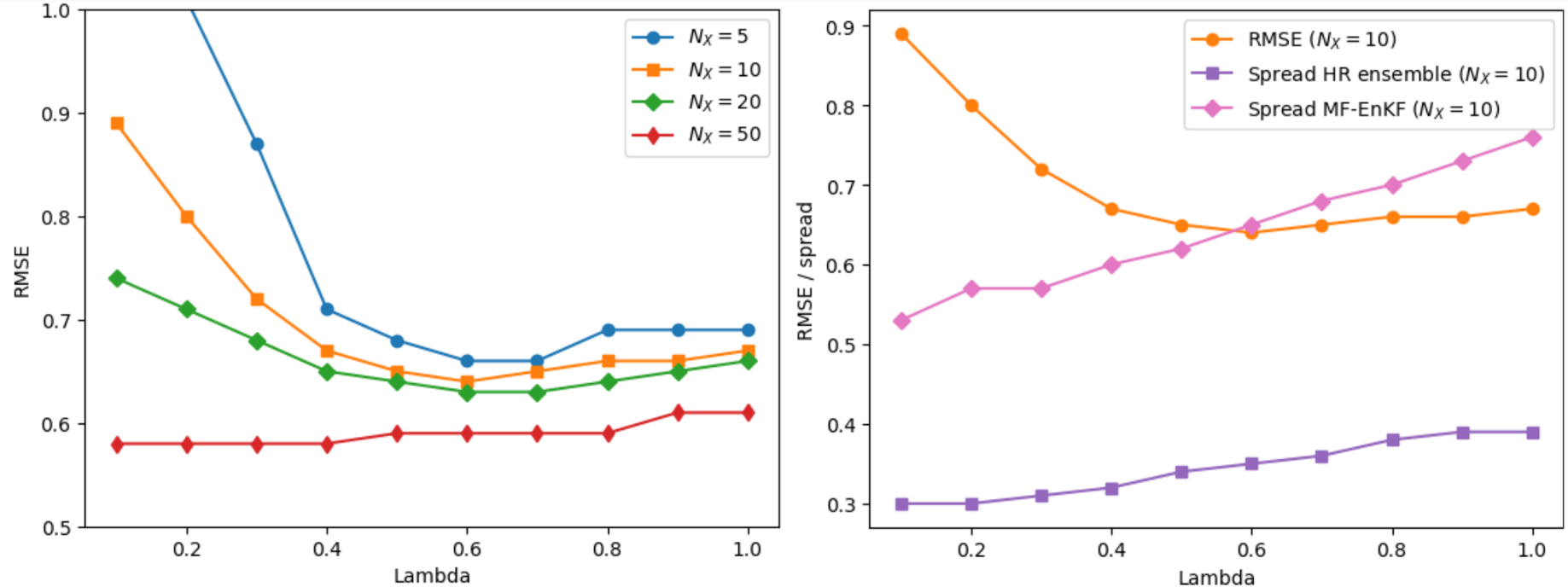}
    \caption{Left: RMSE against different values of the $\lambda$ parameter, $N_U$ is fixed at 100. Right: RMSE and spread of the HR ensemble and MF-EnKF for $N_X = 10$ and $N_U = 100$}
    \label{fig:infl_of_lambda_QG}
\end{figure}

Again, we check the sensitivity of the MF-EnKF results to the tuning parameter $\lambda$, which we have fixed at $\lambda = 0.5$ until now. For a few different combinations of $(N_X,N_U)$, we test the influence of $\lambda$ on the RMSE and on the spread of the ensemble, similar to Figure 8 in \cite{barthelemy2024}. The result is shown in Figure \ref{fig:infl_of_lambda_QG}. We can observe that the performance is stable for quite a broad range of $\lambda$ values and the chosen value of $\lambda = 0.5$ is close to optimal in both cases. However, $\lambda$ has a big influence on the spread of the ensemble. In the ideal case, the spread should match the actual RMSE, which is the case for around $\lambda = 0.6$. Furthermore, we can observe that when we underestimate the error we are getting bigger errors than when we overestimate the error. Also, the spread of the HR ensemble is always underestimating the true error, which shows that adding the surrogate runs improves not only the RMSE, but also the estimation of the posterior variance.

\section{Conclusion and discussion} \label{conclusion}

\textbf{Conclusion}. \\
We described a hybrid modelling approach in which a MF-EnKF is combined with a ML surrogate that is trained to emulate the full forward operator, which traditionally has been done using a LR or ROM surrogate. Our method combines an ensemble of a few expensive HR runs, together with many cheap and less accurate ML runs. We compared the performance of our approach with the standard EnKF that includes optimal localization and inflation. Inside the MF-EnKF we chose the deterministic EnKF for convenience, but we expect similar results for other EnKF variants. We considered two test cases: the Lorenz-2005 and the Quasi-Geostrophic model. In both cases, the MF-EnKF with ML surrogate reached improved accuracy within the same computational budget compared to the EnKF. This would be a big advantage in practice, as the computational cost is one of the most limiting factors for ensemble-based data assimilation methods. The optimal performance was achieved for a combination of HR and ML runs, instead of only ML or only HR runs. With this optimal combination, we would effectively be able to run an EnKF with a much larger ensemble size than before, but within the same computational time. \\
\ \\
Furthermore, we showed that using an ML surrogate inside the MF-EnKF resulted in similar or improved accuracy compared to the traditional MF-EnKF, which uses a LR or ROM surrogate, for the same division of $(N_X,N_U)$ ensemble sizes. Although we studied only one particular ML model for both cases, the comparison to LR models indicates that performance of the MF-EnKF behaves as expected and will likely improve with increasing accuracy of the ML model. Furthermore, we expect the LR model to be more expensive than the ML model, since it is limited by the grid size. For instance, reducing the resolution of a 2-dimensional model by a factor of 2 would give an approximately 8 times faster model: a factor 2 in both x- and y-direction and a factor 2 for coarser time steps while still satisfying the CFL condition. On the other hand, the ML model is not constrained by the grid size and able to use the GPU to speed up computations much more. \\
\ \\
Regarding the division of the computational resources, in both of our test cases it seemed that the optimal mix is roughly 50$\%$: spend half of the  resources on the HR model and the other half on the ML model runs. In general, the RMSE curves were quite flat for a large range of combinations of $(N_X,N_U)$ pairs, indicating stable behavior when we have some combination of HR and ML runs, and unstable behavior when we only have HR or only ML runs within the same budget. \\
\ \\
Compared to other approaches, such as the Multi-Level EnKF, the MF-EnKF has an extra tuning parameter $\lambda$. We showed that the MF-EnKF is stable for a large range of $\lambda$ values and $(N_X, N_U)$ pairs, which indicates that the algorithm does not need extensive tuning. Alternatively, one could try to adaptively estimate the $\lambda$ parameter, similar to what is done in \cite{gharamti2020} for a hybrid ensemble-variational filter. For instance, in the case of extreme events one may want to decrease the weight to the ML ensemble since ML models usually have trouble modeling these rare events. \\ 
\ \\
Besides the RMSE performance, we further showed that also the spread of the ensemble is improved when using a Multi-Fidelity approach. This means that the MF-EnKF can give better uncertainty quantification than the EnKF alone, since the quality for the uncertainty quantification in the EnKF is limited because of the small affordable ensemble size. Moreover, a hyper-parameter value of $\lambda$ around 0.5 seemed to give a good agreement between the ensemble spread and the true error. \\
\ \\
\textbf{Discussion}. \\
The main disadvantage of our MF-EnKF variant is that an ML surrogate model should be available. It can take considerable computational time to create enough training data to train this surrogate. Next to that, choosing the right neural network architecture and hyperparameters can be a long process and non-trivial if the network has to satisfy certain physical constraints. However, in applications like weather forecasting, the physical model rarely changes, so the same trained model can be used for a long time and will be worth the training effort. On the other hand, in some applications it can also be hard to set up a LR model, for instance with unstructured grids, and the LR model may be inaccurate due to representation errors or unresolved processes. The choice of surrogate will in the end depend on the application area. \\
\ \\
Even when an ML surrogate is available, making a fair comparison between computational times can be hard. For instance, the computational times depend on the specific implementation, availability of CPU versus GPU, and parallel computing options. In the literature huge speed-up factors are claimed (see Table \ref{tab:ml_weather}). However, if the full model has been implemented on GPU, as with our QG model, then the ML surrogate may not be much faster than the original model. Next to that, depending on the implementation, it may not be easy to connect the ML model code (which is usually in TensorFlow or PyTorch) to the original EnKF implementation code that is being used. \\
\ \\
Finally, in order to obtain stable results with the MF-EnKF, we needed a few heuristic ensemble corrections. Although these are effective, further research is needed into the MF-EnKF to avoid these. For instance, \cite{maurais2023} use Log-Euclidean geometry to estimate the multi-fidelity covariance matrix, to guarantee its positive definiteness. \\
\ \\
\textbf{Recommendations}. \\
Our approach has been shown to work in two simple test cases. The next step would be to apply it to a more realistic test case. If there is a ML model and an EnKF implementation available, then it should be straightforward to apply the MF-EnKF framework. Furthermore, our results suggest that no extensive tuning is needed for the hyper-parameter $\lambda$ and the choice of distributing computational budget between the $N_X$ and $N_U$ members. A particularly interesting application area would be the field of numerical weather prediction (NWP), since there are already several well-developed ML surrogates available in this field (e.g., \cite{pangu}; \cite{fourcastnet}; \cite{aifs}; \cite{aardvark}; \cite{graphcast}; \cite{aurora}). But also for other application areas, ML surrogates have been developed, such as groundwater \cite{groundwater_ML}, hydrology \cite{hydrology_ML} and storm surges \cite{storm_surge_ML}. Our method allows practitioners to generate a large number of ensemble members, which has not been possible before, improving estimation of the initial condition and hence accuracy of the predictions in fields such as meteorology, oceanography and other fields in which ensemble data assimilation is commonly used.

\bibliographystyle{unsrt}
\bibliography{references.bib}

\appendix \label{app:QG_param}

\section*{Appendix. Derivation of QG parameters}

In this Appendix, we will derive the values for the QG model parameters based on the values in \cite{sakov2008}, which in turn are based on the QG model as given in \cite{jelloul2003}. We start from the 3-layer QG model from \cite{thiry2022} for variables pressure ($p$) and potential vorticity ($q$):
\begin{equation} \label{QG_Thiry}
    \begin{cases}
        q_t = \frac{1}{f_0} J(q,p) + f_0 B e + \frac{1}{f_0} a_2 \Delta^2 p - \frac{1}{f_0} a_4 \Delta^3 p, \\
        \Delta p - f_0^2 A p = f_0 q - f_0 \beta (y - y_0)
    \end{cases}
\end{equation}
where
$$ B = \begin{bmatrix} \frac{1}{H_1} & \frac{-1}{H_1} & 0 & 0 \\ 0 & \frac{1}{H_2} & \frac{-1}{H_2} & 0 \\ 0 & 0 & \frac{1}{H_3} & \frac{-1}{H_3} \end{bmatrix}, \qquad e = \begin{bmatrix} \partial_x \tau_2 - \partial_y \tau_1 \\ 0 \\ 0 \\ \frac{ \delta_{ek}}{2 |f_0|} \Delta p_3 \end{bmatrix}, \qquad \tau = \tau_0 \begin{bmatrix} - \cos(2 \pi y / L_y) \\ 0 \end{bmatrix}, \qquad A = \begin{bmatrix} \frac{1}{H_1 g'} & \frac{-1}{H_1 g'} & 0 \\ \frac{-1}{H_2 g'} & \frac{1}{H_2} \left( \frac{1}{g'} + \frac{1}{g''} \right) & \frac{-1}{H_2 g''} \\ 0 & \frac{-1}{H_3 g''} & \frac{1}{H_3 g''} \end{bmatrix}. $$
We want to rewrite \eqref{QG_Thiry} in terms of the streamfunction, as given in Equation (1) from \cite{jelloul2003}. First note that, due to geostrophic balance, we have
$$  f_0 u = -p_y, \qquad   f_0 v = p_x. $$
Using that the streamfunction $\psi$ is defined as $u = -\psi_y$ and $v = \psi_x$, we can express the pressure $p$ in terms of $\psi$,
$$ f_0 \psi_y = p_y, \qquad 
    f_0 \psi_x = p_x. $$
Substituting this into \eqref{QG_Thiry} and rescaling the forcing by $\frac{1}{f_0}$ (as is done in the implementation of \cite{thiry2022}), we find
\begin{equation*}
    \begin{cases}
        q_t = J(q,\psi) + B e + a_2 \Delta^2 \psi - a_4 \Delta^3 \psi \\
        \Delta \psi - f_0^2 A \psi = q - \beta(y-y_0)
    \end{cases}
\end{equation*}
We can substitute $q = \Delta \psi - f_0^2 A \psi + \beta(y-y_0)$ into the first equation to obtain
$$ q_t = \partial_t (\Delta \psi - f_0^2 A \psi + \beta(y-y_0) ) = \partial_t (\Delta \psi - f_0^2 A \psi), $$
with the Jacobi operator equal to
$$ J(q, \psi) = q_x \psi_y - \psi_x q_y = (\Delta \psi - f_0^2 A \psi + \beta (y - y_0) )_x \psi_y - \psi_x (\Delta \psi - f_0^2 A \psi + \beta (y - y_0) )_y $$
$$ = ( (\Delta \psi)_x - f_0^2 A \psi_x) \psi_y - \psi_x ( (\Delta \psi)_y - f_0^2 A \psi_y + \beta ) $$
$$ = (\Delta \psi)_x \psi_y - \psi_x (\Delta \psi)_y + f_0^2 A (-\psi_x \psi_y + \psi_x \psi_y) - \beta \psi_x $$
$$ = - \beta \psi_x + J(\Delta \psi, \psi). $$
So we find
\begin{equation} \label{eq:QGthiry_psi_min}
    \partial_t (\Delta \psi - f_0^2 A \psi) = -\beta \psi_x + J(\Delta \psi, \psi) + B e + a_2 \Delta^2 \psi - a_4 \Delta^3 \psi
\end{equation}
The model from \cite{jelloul2003} is a reduced-gravity model, which is equal to a multi-layer QG model with a second layer that has infinite depth. Hence, letting $H_2, H_3 \rightarrow \infty$ (so that also potential vorticity $q_2 = q_3 = 0$), we have
$$ f_0^2 A = \frac{f_0^2}{H g'}, \qquad f_0 B e = \frac{1}{H} \tau_0 \frac{2 \pi}{L} \sin \left( \frac{2 \pi y}{L} \right ). $$
So that the system of equations \eqref{eq:QGthiry_psi_min} reduces to the single equation
\begin{equation} \label{eq:QGthiry_final}
    \partial_t \left(\Delta \psi - \frac{f_0^2}{H g'} \psi \right) = -\beta \psi_x + J(\Delta \psi, \psi) + \frac{ 2 \pi \tau_0 }{H L} \sin \left( \frac{ 2 \pi y }{L} \right) + a_2 \Delta^2 \psi - a_4 \Delta^3 \psi.
\end{equation}
We assume no horizontal friction, hence $a_2 = 0$. To arrive at the dimensionless QG model from \cite{sakov2008}, we apply the following adimensionalization procedure:
$$ (x,y) \rightarrow L (\tilde{x}, \tilde{y} ), \qquad t \rightarrow \frac{1}{\beta L} \tilde{t}, \qquad \psi \rightarrow \frac{\tau_0}{\rho_0 H \beta} \tilde{\psi}, $$
where the tilde denotes the dimensionless variable. Note that \cite{thiry2022} express $\tau_0$ in $m^2/ s^{2}$ (instead of $N / m^2$), which is already scaled by $\rho_0 = 1000$ $kg/m^3$, so we can remove $\rho_0$ from the equation. In this way we can express the partial derivatives as follows:
$$ \frac{\partial }{ \partial t} = \beta L \frac{ \partial }{ \partial \tilde{t} }, \qquad  \frac{\partial}{\partial x} = \frac{1}{L} \frac{ \partial }{\partial \tilde{x} }, \qquad  \Delta _{x,y} = \frac{ \partial^2 }{ \partial x^2 } + \frac{ \partial^2}{\partial y^2} = \frac{1}{L^2} \Delta_{\tilde{x},\tilde{y} }. $$
So we obtain
$$ \frac{ \partial \psi}{\partial x} = \frac{ \tau_0}{ H \beta} \frac{1}{L} \frac{ \partial \tilde{\psi}}{\partial \tilde{x} }, 
\quad \frac{ \partial \psi}{\partial t} = \beta L \frac{ \tau_0}{ H \beta} \frac{ \partial \tilde{\psi}}{\partial \tilde{t} }, 
\quad \Delta_x \psi = \frac{\tau_0}{ H \beta} \frac{1}{L^2} \Delta_{\tilde{x}} \tilde{\psi}, 
\quad \Delta_x^3 \psi = \frac{ \tau_0 }{ H \beta} \frac{1}{L^6} \Delta_{\tilde{x} }^3 \tilde{\psi}. $$
Hence, \eqref{eq:QGthiry_final} in terms of $\tilde{\psi}, \tilde{x}, \tilde{y}$ and $\tilde{t}$ is
$$ \beta \frac{\tau_0}{ H \beta} \frac{1}{L} \partial_{\tilde{t} } \left( \Delta \tilde{\psi} - R_d^{-2} L^2  \tilde{\psi} \right) = - \beta \frac{\tau_0}{ H \beta} \frac{1}{L} \partial_{\tilde{x}} \tilde{\psi} + \frac{1}{L^2} \left( \frac{\tau_0}{H \beta } \right)^2 \frac{1}{L^2} J( \Delta \tilde{\psi}, \tilde{\psi}) + \frac{2 \pi \tau_0}{H L} \sin ( 2 \pi \tilde{y} ) - a_4 \frac{\tau_0 }{ H \beta} \frac{1}{L^6} \Delta^3 \tilde{\psi} $$
This simplifies to
$$ \partial_{\tilde{t} } \left( \Delta \tilde{\psi} - R_d^{-2} L^2  \tilde{\psi} \right) = -\partial_{\tilde{x}} \tilde{\psi} + \frac{ \tau_0}{ H \beta^2 L^3} J(\Delta \tilde{\psi}, \tilde{\psi} ) + 2 \pi  \sin( 2 \pi \tilde{y} ) - a_4 \frac{1}{\beta L^5} \Delta^3 \tilde{\psi}. $$
Removing the tilde symbol, we find
$$ q_t = -\psi_x - \varepsilon J(\psi, q) - A \Delta^3 \psi + 2 \pi \sin(2 \pi y), $$
with $q = \Delta \psi - F \psi$ and
$$ F = \frac{f_0^2 L^2}{g' H}, \qquad \varepsilon = \frac{ \tau_0 }{ H \beta^2 L^3}, \qquad A = a_4 \frac{1}{\beta L^5}. $$
Comparing this with Equation (20) from \cite{sakov2008} and their values $F=1600$, $\varepsilon = 10^{-5}$ and $A = 2 \cdot 10^{-12}$, we find the following equations for the parameters of the QG model:
$$ \frac{f_0^2 L_y^2}{g' H} = 1600, \qquad  \frac{ \tau_0}{\rho_0 H \beta^2 L_y^3} = 10^{-5}, \qquad a_4 \frac{1}{\beta L^5} = 2 \cdot 10^{-12} $$
\begin{equation} \label{QGparam_eq}
   \Rightarrow \qquad H = \frac{f_0^2 L^2}{1600 g'}, \qquad \tau_0 = 10^{-5}  H \beta^2 L^3, \qquad  a_4 = 2 \cdot 10^{-12} \beta L^5
\end{equation}
The remaining unknowns are $g', L, f_0$ and $\beta$. We assume reduced gravity $g' = 0.025$, which is the value used in \cite{thiry2022} and \cite{thiry2024}. To find the basin length $L$, we can use that according to \cite{barthelemy2024} the distance between satellite tracks across the basin is 400 km and that they make an angle of approximately 66 degrees. Given that 7 times the horizontal distance between tracks fits into $L$ (see Figure 1 from \cite{sakov2008}), we can find $L$ as
$$ L = \frac{2800 \text{ km}}{\sin(66) } \approx 3065 \text{ km}. $$
Given that $n_x = 129$, we choose $L = 3072$ km such that the resolution is exactly 24 km. For the low-resolution we have $n_x = 65$ and hence the resolution is exactly 48 km. We assume a time step of $\Delta t = 6$ hours. In \cite{sakov2008} a dimensionless time step of 1.25 is used. Since $t = \frac{\tilde{t}}{\beta L}$, we have
$$ \beta = \frac{ \tilde{t}}{t L} \approx 1.88 \cdot 10^{-11} \ m^{-1} s^{-1}. $$
To derive $f_0$, we use the following relationship between $f_0, \beta$ and the latitude $\theta$ (in degrees):
$$ f_0 = 2 \Omega \sin ( \theta \pi / 180), \qquad \beta = 2 \Omega / R_e \cos( \theta \pi / 180), $$
with $\Omega = 7.2921 s^{-1}$ the angular speed of the earth and $R_e = 6371 \cdot 10^3$ m the radius of the earth. Then we can derive that
$$ \theta = \frac{180}{\pi} \arcsin \left( \frac{f_0}{2 \Omega} \right) \quad \Rightarrow \quad \theta \approx 35 \text{ degrees}, $$
from which it follows that $f_0 \approx 8.4 \cdot 10^{-5}$. Now we can finally solve \eqref{QGparam_eq} to obtain
$$ H = 1664 \ m, \qquad \tau_0 = 1.7 \cdot 10^{-4} \ m^2s^{-2}, \qquad a_4 = 1 \cdot 10^{10} \ m^4 s^{-1}. $$
A summary of the parameters and their meanings and final values is given in Table \ref{tab:QGparam} in the main text.

\end{document}